\documentclass{article}
\usepackage{ijcai20}

\usepackage{times}

\usepackage{soul}
\usepackage{url}
\usepackage[hidelinks]{hyperref}
\usepackage[utf8]{inputenc}
\usepackage[small]{caption}
\usepackage{graphicx}
\usepackage{amsmath}
\usepackage{booktabs}
\usepackage{subcaption}
\title{Towards Utilizing Unlabeled Data in Federated Learning: A Survey and Prospective}

\author{
Yilun Jin$^1$\footnote{Contact Author}\and
Xiguang Wei$^2$\and
Yang Liu$^{2}$
\And Qiang Yang$^{1,2}$
\\
\affiliations
$^1$The Hong Kong University of Science and Technology, Hong Kong SAR, China\\
$^2$Webank, Shenzhen, China\\
\emails
yilun.jin@connect.ust.hk,
\{xiguangwei,yangliu\}@webank.com,
qyang@cse.ust.hk
}

\begin{document}

\maketitle

\begin{abstract}
Federated Learning (FL) proposed in recent years has received significant attention from researchers in that it can bring separate data sources together and build machine learning models in a collaborative but private manner. Yet, in most applications of FL, such as keyboard prediction, labeling data requires virtually no additional efforts, which is not generally the case. In reality, acquiring large-scale labeled datasets can be extremely costly, which motivates research works that exploit unlabeled data to help build machine learning models. However, to the best of our knowledge, few existing works aim to utilize unlabeled data to enhance federated learning, which leaves a potentially promising research topic. In this paper, we identify the need to exploit unlabeled data in FL, and survey possible research fields that can contribute to the goal. 
\end{abstract}

\section{Introduction}
\label{sec:intro}
There should be little doubt that the prosperity of \textit{Artificial Intelligence} (AI) should largely be attributed to the availability of Big Data. As an example, the field of computer vision, where we witnessed numerous advances in deep learning, was significantly boosted with the advent of the comprehensive ImageNet dataset \cite{deng2009imagenet}. 

Yet when it comes to applications of AI in real-world scenarios, things are not exactly the case. It is often the case that corporations only possess low-quality, incomplete and insufficient data. To this end, \textit{Federated Learning} \cite{mcmahan2017communication,yang2019federated,kairouz2019advances} was proposed as an attempt to alleviate such a problem by enabling private collaboration among parties without explicit sharing of data. Up till now, FL has been widely accepted as a new learning scheme and has triggered numerous applications \cite{hard2018federated,chen2019federated,yang2018applied}. 

Nonetheless, as we observe the existing applications of FL, we find that the majority of them require no additional efforts to label the data. For example, in next-word prediction \cite{hard2018federated}, data are automatically labeled through user typing behaviors. Yet in general, raw data collected require manual labeling, which makes it hard to obtain large-scale, high-quality labeled datasets, making the application of FL limited. 
\begin{table*}[ht]
    \centering
    \begin{tabular}{|c|c|c|c|}
        \hline
        \textbf{FL setting} & \textbf{ID Space} & \textbf{Feature Space} & \textbf{Label Space}\\
        \hline
        Horizontal Federated Learning (HFL) & Different & Same & Same \\
        \hline
        Vertical Federated Learning (VFL) & Same/Can be aligned & Different & Different\\
        \hline
        Federated Transfer Learning (FTL) & Different & (Generally) Different & (Generally) Different\\
        \hline
    \end{tabular}
    \caption{FL Categorization According to Data Partition}
    \label{tab:flcat_partition}
\end{table*}

\begin{table*}[ht]
    \centering
    \begin{tabular}{|c|c|c|c|c|}
        \hline
        \textbf{FL setting} & \textbf{Participants} & \textbf{\# Participants} & \textbf{Local Dataset Size} & \textbf{Consistency}\\
        \hline
        Cross-device FL & e.g. phones, IoT devices. & Massive, up to $10^{10}$ clients & Relatively small & Inconsistent\\
        \hline
        Cross-silo FL & e.g. corporations, institutes & Up to $10^2$. & Relatively large & Consistent\\
        \hline
        
    \end{tabular}
    \caption{FL Categorization According to Type of Participants. The term 'consistency' means the consistency of participants across each round. In cross-device FL, the participants are not always available (e.g. subject to network and battery status, and diurnal-nocturnal changes), making the participants for each round different, and thus 'inconsistent'. On the contrary, cross-silo FL shows much better consistency, as they use dedicated hardware, reliable networks and are much better scheduled. }
    \label{tab:flcat_type}
\end{table*}

\begin{figure*}[ht]
    \centering
    \includegraphics[width=0.8\linewidth]{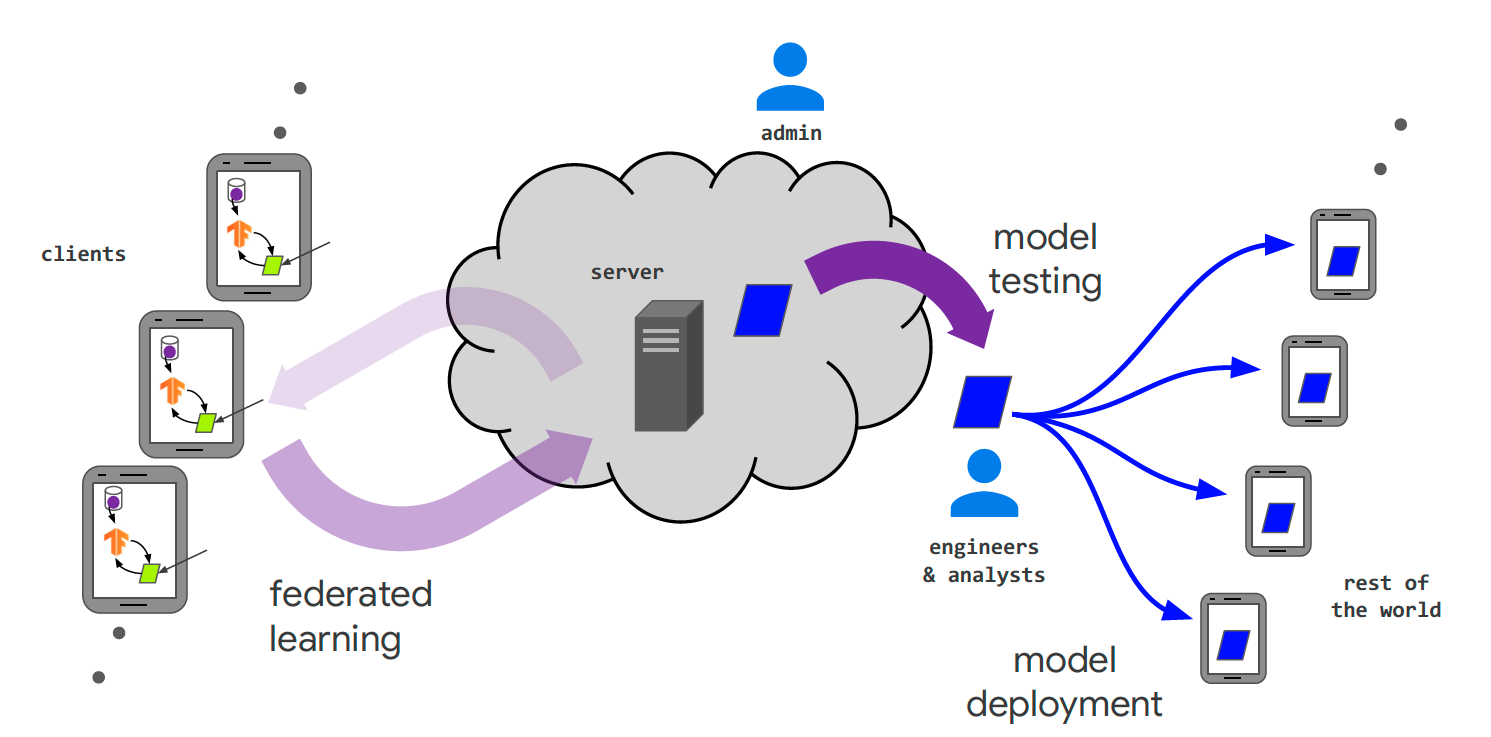}
    \caption{An illustration of cross-device FL. The model is trained through numerous devices and is deployed to all devices throughout the world. }
    \label{fig:cross-device-fl}
\end{figure*}
We argue that applications of FL are in more pressing need of utilizing unlabeled data than others. On one hand, in cross-device FL \cite{kairouz2019advances}, where participants are individual devices, numerous unlabeled data are generated through our interaction with smart devices, such as photos taken, text inputs, and physiological indicators measured by wearables, whose sheer volume makes it impractical to require users to label them. On the other hand, in cross-silo FL where participants are corporations, the data involved are likely to require human expertise, such as finance (risk management, credit evaluation), and medical applications (disease diagnosis, health monitoring). In these cases, it would require significant human intellect and efforts to label the data. In this case, labeling all the data would be costly, and thus makes it necessary to utilize unlabeled data and learn models in a weakly supervised manner. 

Nevertheless, compared to other areas, there is relatively little attention paid to this area. While techniques like transfer learning, semi-supervised learning, self-supervised learning and active learning are all popular research topics, we can only observe popularity in federated transfer learning (FTL) \cite{Peng2020Federated,liu2018secure}, while others are relatively ignored. 

Consequently, in this paper, we seek to provide a perspective into weakly supervised approaches in federated learning. We first introduce related preliminaries, before identifying motivations that drive us to devote to this problem. Last but not least, we make a prospect into potential scenarios, research topics, as well as challenges. We hope that our efforts can be followed by researchers who come up with concrete solutions to the problem that will contribute to both the academia and the industry. 

% \begin{figure*}
% \centering
% \includegraphics[width=0.7\textwidth]{HFL.png}
% \caption{Architecture For Horizontal Federated Learning}
% \label{fig:HFL}
% \end{figure*}

\section{Preliminaries and Related Work}
\label{sec:prelim}
\subsection{Federated Learning}
\textit{Federated Learning}, proposed by \cite{mcmahan2017communication} and extensively surveyed by \cite{yang2019federated,kairouz2019advances}, is a machine learning scheme that enables aggregation of isolated data in a privacy-preserving manner. Generally speaking there are two major categorization standards proposed by previous surveys, with the first \cite{yang2019federated} focusing on data partitions and the latter \cite{kairouz2019advances} focusing on types of participants. We show the two categorizations in Table \ref{tab:flcat_partition} and Table \ref{tab:flcat_type}, respectively.

%\begin{enumerate}
%    \item \textbf{Horizontal federated learning (HFL)}, which means that datasets held by clients share the same feature and label space but different sample id spaces, i.e. $$\mathcal{I}_j \neq \mathcal{I}_k, \mathcal{X}_j = \mathcal{X}_k, \mathcal{Y}_j = \mathcal{Y}_k, \forall k\neq j$$
%    where $k$ and $j$ are different clients. 
%    \item \textbf{Vertical federated learning (VFL)}, which means that clients share the same sample id space yet different feature spaces. Also, the label spaces of different clients may be different (as illustrated by certain clients having labels, while others not),  i.e. $$\mathcal{I}_j =  \mathcal{I}_k, \mathcal{X}_j \neq \mathcal{X}_k, \mathcal{Y}_j \neq \mathcal{Y}_k, \forall k\neq j.$$
%    \item \textbf{Federated transfer learning (FTL)} \cite{liu2018secure} describes a more general and challenging scenario where the clients all hold different feature spaces, label spaces and sample id spaces, i.e. 
    %$$\mathcal{I}_j \neq \mathcal{I}_k, \mathcal{X}_j \neq \mathcal{X}_k, \mathcal{Y}_j \neq \mathcal{Y}_k, \forall k\neq j.$$
%\end{enumerate}

Existing works on FL have shown significant diversity. There have been research works on federated optimization \cite{mcmahan2017communication,Li2020On,Wang2020Federated}, federated learning algorithms \cite{cheng2019secureboost,li2019practical,shokri2015privacy}, privacy mechanisms and attacks \cite{hitaj2017deep,mohassel2017secureml,bonawitz2017practical}, systems and communication \cite{bonawitz2019towards}, etc. However, regarding FL in weakly-supervised scenarios, relatively little attention has been paid to this area. 
%It is worth mentioning that data privacy should be maintained throughout the learning process, as stated in \cite{yang2019federated}. Generally the privacy requirement rules out not only explicit data transfer, but also more intricate forms of privacy leakage, e.g. model inversion attacks \cite{fredrikson2015model}, membership inference attacks \cite{shokri2017membership}, etc. While these forms of attacks do pose threats to federated semi-supervised learning, they focus on federated learning as a whole rather than specifically on federated semi-supervised learning. Consequently, for simplicity, we only focus on privacy issues induced by semi-supervised training in this paper, instead of general privacy issues and defenses. 

Existing works on weakly supervised FL mostly fall into federated transfer learning (FTL), with \cite{Peng2020Federated} and \cite{liu2018secure} proposed unsupervised and supervised FTL, respectively. There are also works tackling federated self-supervised feature learning on texts \cite{jiang2019federated,mcmahan2017communication} by learning topic models and language models. Regarding other forms of weakly supervised algorithms, such as semi-supervised learning and active learning, we observe little prior arts \cite{goetz2019active} to the best of our knowledge. 

We here discuss two prior works on federated transfer learning. \cite{liu2018secure} tackles the problem of semi-supervised transfer learning between two clients, where the two clients exchange gradients and intermediate results through Homomorphic Encryption (HE). As generally, HE is computationally expensive to perform, the approach may not scale to cross-device FL where maybe millions of participants exist. \cite{Peng2020Federated} focuses on unsupervised domain adaptation, that uses several source domains held by clients to facilitate classification on one target domain. The work achieves domain adaptation through novel adversarial training techniques and achieved convincing results. Yet, similar to \cite{liu2018secure}, this work assumes that the participants are static and constantly available, which also does not scale to the cross-device FL setting. 
\subsection{Weakly Supervised Learning Algorithms}
\subsubsection{Transfer Learning}
Transfer Learning \cite{yang2020transfer} aims to transfer knowledge learned from a source domain to a relevant target domain, probably with fewer labeled samples to train on. Existing popular transfer learning methods include domain adaptation \cite{long2014domain,long2015learning}, knowledge distillation \cite{hinton2015distilling}, and pre-training/fine-tuning \cite{devlin2019bert} etc. 

While transfer learning has achieved tremendous success in vision and language modeling, and even triggered interests in FTL, one limitation exists, that a related source domain with abundant data must be found to support transfer learning. In FL, the applications are highly diverse, which makes it hard for every one of them to find a suitable and resourceful source domain. 

\subsubsection{Semi-supervised Learning}
Semi-supervised Learning (SSL) \cite{zhu2005semi} aims to learn a model under very limited labeled data and also massive unlabeled data. SSL is widely adopted in areas where labels are scarce. In most cases researchers utilize unlabeled data to improve the generalization performance and prevent overfitting caused by small datasets. Popular methods of SSL include generative models \cite{kingma2014semi,robert2018hybridnet}, adversarial training \cite{miyato2018virtual,odena2016semi}, regularization \cite{tarvainen2017mean}, pseudo-labeling \cite{berthelot2019mixmatch}, connections between samples \cite{kipf2016semi} and multi-view ensemble training \cite{chen2018tri}. 

\subsubsection{Self-supervised Learning}
Self-supervised learning, also known as representation learning, aims to extract indicative features from large amounts of data without label supervision. Consequently, common approaches in self-supervised learning utilize the data themselves to provide supervision, trying to capture innate structures within the data. Up till now, self-supervised learning has achieved tremendous success in natural language process (NLP) through large-scale language models \cite{devlin2019bert}, and also topic models \cite{jiang2019federated}. Also in the area of vision, self-supervised feature learning is popular, commonly achieved by learning colorization, positioning and rotation information \cite{trinh2019selfie,pathak2016context}, and has been used for boosting performance in semantic segmentation, clustering, and object detection \cite{jing2019self}.

\subsubsection{Active Learning}
Active Learning aims to train a classifier on datasets with few labeled samples by making as few queries of additional label information as possible. Essentially, active learning aims to find samples that, when labeled, will provide the greatest contribution towards model learning. In existing approaches, active learning is achieved by designing label query algorithms, such as the most uncertain samples \cite{settles2008analysis}, most variance reduction \cite{schein2007active}, etc. 

\section{Motivations and Advantages}
In this section, we identify the motivations that drive us to the problem of FL in weakly supervised settings, and propose advantages that will arise when FL is able to utilize unlabeled samples. 

\subsection{Expanding Application Scenarios}
Existing application scenarios of FL generally work on problems which require little extra effort to label the data. For example, in language modeling \cite{mcmahan2017communication}, labeling is automatically achieved through user typing behaviors. In recommendation \cite{qiang2019federated}, the labels are purchase records of users, which also require no extra labor. Yet in most applications, explicit labeling is required, such as object recognition, sentiment analysis, person re-identification, etc. 

We also argue that applications of FL face even greater demands in utilizing unlabeled data. 
\begin{itemize}
    \item First, FL imposes strong privacy requirements, which rules out large-scale labeling through outsourcing, which is a common practice in corporations.
    \item Second, in cross-device FL introduced by \cite{kairouz2019advances}, where participants are smart devices, huge amounts of data are generated every day, such as text inputs, images taken, and even physiological indicators measured by wearables. These data are either too large in size to require users to label, or require high-level human expertise (such as sleep monitoring, heartbeats) that few users possess. Consequently, quite often the data generated remain unlabeled. 
    \item Last but not least, in cross-silo FL, where participants are corporations, the data involved often lie within specialized domains, such as finance (risk management, credit evaluation, anti money laundering), or clinical services (medical image diagnosis, object detection and localization). In these domains, the effort required to label the data are generally prohibitive, and therefore we can only afford to label a small proportion of them, instead of the whole dataset. 
\end{itemize} 

Consequently, developing algorithms that effectively utilize unlabeled data to enhance training would open up extensive new applications and help build a more vibrant federated AI ecosystem. 

\subsection{Mitigating Domain Discrepancy}
As a challenge identified by many researchers, non-iid data is a prominent issue in FL, and there have also been works to study such a challenge \cite{Li2020On}. Generally speaking, non-iid data pose two challenges to FL. On one hand, the data owned by different parties inevitably differ in their distribution, causing difficulties in model learning. On the other hand, domain discrepancy also exists between training and testing. Chances are that the data used to train a federated model differs a lot to those owned by certain users, making the model ineffective for them. In fact, a recent empirical study \cite{yu2020salvaging} demonstrated that, federated language models can be less accurate than a considerable proportion (as much as 20\%) of local models trained using data from individual parties, whose data distributions differ a lot from the global distribution. 

Utilizing large-scale unlabeled data, correspondingly, is able to mitigate the problem of non-iid data. Intuitively, by viewing a sufficiently large unlabeled dataset, one can get a much better understanding of the data distribution than using only a small labeled dataset alone. For example, unlabeled data can be used to train generative models that provide additional information about the data's prior distribution $p(x)$, thus filtering out the domain-specific features \cite{kingma2014semi,robert2018hybridnet}. In addition, domain adaptation that minimizes domain discrepancies can also be used on unlabeled data \cite{Peng2020Federated}, such that domain invariant representations can be learned. Last but not least, advances in disentangled representations \cite{siddharth2017learning} can also contribute to domain invariant models by disentangling domain-specific features from domain invariant ones. 

\subsection{Enhancing Robustness}
Robustness means that a model would be resilient to small variations, such as outliers and small perturbations of inputs, which is appealing in most machine learning applications. By utilizing unlabeled data to regularize the model, robustness can be achieved. For example, sensitivity towards small perturbations can be alleviated if we regularize the model to produce consistent outputs in the neighborhood of each data point. It would not be possible if only a few labeled samples are available, as they only represent a small subset over the data distribution. In addition, reliance on specific data points can be alleviated if more unlabeled data can be used to prevent overfitting on a few labeled samples. 

Robustness in FL also implies attractive outcomes. On one hand, when participants of FL have a rather limited amount of data, the local trainings are likely to be noisy, and local models prone to overfitting. By utilizing available unlabeled data for regularization, local overfitting can be alleviated and therefore, a better global model can be reached. On the other hand, robustness implies resilience towards modification of the dataset, which is favorable towards private and secure machine learning models. For example, robustness against small perturbations would lead to resistance over data poisoning attacks, such as adversarial examples \cite{goodfellow2014explaining}. As another example, as shown in \cite{shokri2017membership}, membership inference attacks are closely related to overfitting, and the more overfitting the model is, the more prone it is towards membership inference attacks (as the model is more likely to behave differently on samples that are used to train the model). Consequently, robustness in FL can also lead to appealing properties in security. 

\section{Potential Topics and Challenges}
In this section we introduce potential settings and topics, both in research and applications, that may contribute to better FL algorithms, and also potential challenges that may arise. 
\subsection{Transfer Learning}
Existing solutions enabling FTL have been highly sophisticated \cite{liu2018secure,Peng2020Federated}. We here identify several potential topics regarding FTL. 
\begin{itemize}
    \item \textbf{Versatile Source Domains and Datasets.} As FL should support a wide range of applications, to enable FTL, it is important that adequate source domains and datasets are chosen, otherwise negative transfer \cite{cao2010adaptive} may happen. It is thus important in practice that adequate source domains must be chosen to enable FTL applications. Alternatively, it is always welcomed to develop versatile datasets that transfer to multiple domains. 
    \item \textbf{Realistic Federated Datasets.} FL features non-iid data held by different participants, as determined by location, population, etc, and a realistic federated dataset that accurately replicates such domain discrepancies would be necessary for evaluating FTL or even broader FL algorithms. Up till now, existing FTL evaluations use artificial datasets created by manipulating existing benchmarks, which may not accurately capture real-world domain discrepancies featured by FL. 

    \item \textbf{FTL in cross-device FL.} Existing solutions on FTL work on relatively few participants, e.g. several, or tens, with each of them holding relatively large data, and are always available throughout the training \cite{Peng2020Federated}. Yet, in cross-device FL, participants are much larger in size, inconsistent for each round of training, and each of them may hold much smaller amounts of data, as shown in Table \ref{tab:flcat_type}. It is thus relatively unknown how FTL can work in the cross-device FL setting, which also shows significant domain discrepancy \cite{yu2020salvaging}.
\end{itemize}
\subsection{Semi-supervised Learning}
Semi-supervised setting in FL has received little attention, which leaves a promising potential topic, as semi-supervised learning can work on almost all types of data. For example, in medical image classification, obtaining fully annotated training datasets may not be possible, where we can resort to federated semi-supervised learning to solve the problem. As another example, it is also costly to obtain fully annotated data in financial applications, where collaborators such as banks, insurance companies would jointly train their model in a semi-supervised manner. We here point out several potential challenges that need to be resolved in this topic. 
\begin{itemize}
    \item \textbf{Privacy Requirements. } In certain semi-supervised learning algorithms, connections between samples are leveraged to infer or 'propagate' labels towards unlabeled samples \cite{kipf2016semi}. In these approaches, it is important that privacy requirements are not breached when we leverage these connections. There are also algorithms that involve generative models, which are capable of generating artificial samples \cite{robert2018hybridnet,springenberg2015unsupervised}. Whether such artificial samples are breaking the privacy requirements remains an important challenge that is yet to be resolved. 
    \item \textbf{Domain Discrepancy} Non-iid data always pose significant challenges in FL. In the case of semi-supervised learning, \cite{NIPS2018_7585} showed that when labeled data and unlabeled data belong to different domains (i.e. domains that show significant discrepancy), semi-supervised learning algorithms will significantly degrade in performance. Thus, semi-supervised learning methods in FL must be combined with techniques that tackle with domain discrepancy. 
    \item \textbf{Extension to VFL} Existing studies on semi-supervised learning mainly fits in with the HFL setting, where the unlabeled data are shown intact. However, when it comes to VFL, where the data samples themselves are fragmented and cannot be brought together, more sophisticated protocols should be designed. 
    \item \textbf{Relationship between robustness and security} As mentioned before, model robustness (e.g. robustness to perturbations, outliers) are intuitively related with defense against attacks, such as adversarial attacks and membership inference attacks. As various regularization techniques are involved in semi-supervised learning, it is interesting to study, both empirically and theoretically how such regularization and robustness will contribute towards model security. 
\end{itemize}
\subsection{Self-supervised Learning}
One significant doubt on self-supervised learning in FL is that, it may depend strongly on the data domain and the downstream task it is used for. For example, while self-supervised language modeling is competitive in a wide range of tasks, self-supervised learning in vision is not the case. As shown in \cite{goyal2019scaling}, self-supervised learning is competitive in object detection, but outperformed by supervised pre-training significantly in various classification tasks. Consequently, although self-supervised learning is a natural idea in FL \cite{mcmahan2017communication}, whether it enables wider application may be doubtful and depend heavily on the specific application. 
\subsection{Active Learning}
Active learning seems a natural idea that can be well combined with FL. For example, in cross-device FL, the model holder may ask certain users to label several examples which are then used for training, acting in a crowd-sourcing manner. In cross-silo FL, an institute may identify several difficult examples during training, and ask its experts to label it to facilitate training. 

A key challenge that needs to be solved is how to identify data samples that contribute most to training and should be queried. In federated learning, neither the coordinator or the training server can directly observe raw data. Instead they can only observe batched, and in some cases even protected (e.g. via differential privacy) or encrypted intermediate results. Consequently, identifying individual data samples that may contribute most to training is not straightforward. 

\section{Conclusion}
In this paper we identify a potentially important topic in federated learning: utilizing unlabeled data for weakly supervised federated training. We introduce existing methods that effectively leverage unlabeled data for training models, and point out motivating advantages that arise if unlabeled data can be incorporated for weakly-supervised training. Finally, we make a prospect into potential topics, application scenarios and challenges that come along weakly supervised learning in FL. We hope that this paper can lead to more attempts in more effective utilization of data, better learning algorithms, and a more diverse federated ecosystem featuring a wider range of applications. 

\bibliographystyle{named}
\bibliography{ijcai20}

\end{document}